\documentclass{article}

\usepackage[final]{corl_2018} 

\usepackage[utf8]{inputenc} 
\usepackage[T1]{fontenc}    
\usepackage{hyperref}       
\usepackage{url}            
\usepackage{booktabs}       
\usepackage{amsfonts}       
\usepackage{nicefrac}       
\usepackage{microtype}      
\usepackage{graphicx}       
\usepackage{grffile}   
\usepackage{subcaption}    
\usepackage{amsmath}  
\usepackage{paralist}
\usepackage{verbatim}   
\usepackage[ruled, vlined, linesnumbered]{algorithm2e}
\newenvironment{algorithmicarchitecture}[1][htb]
  {
   \begin{algorithm}[#1]%
  }{\end{algorithm}}

\title{Curiosity Driven Exploration of Learned Disentangled Goal Spaces}

\author{
  Adrien Laversanne-Finot\\
  Flowers Team\\
  Inria and Ensta-ParisTech, France\\
  \texttt{adrien.laversanne-finot@inria.fr} \\
   \And
   Alexandre Péré \\
   Flowers Team \\
   Inria and Ensta-ParisTech, France\\
   \texttt{alexandre.pere@inria.fr} \\
   \AND
   Pierre-Yves Oudeyer \\
   Flowers Team \\
   Inria and Ensta-ParisTech, France\\
   \texttt{pierre-yves.oudeyer@inria.fr} \\
}

\begin{document}
\maketitle


\begin{abstract}
Intrinsically motivated goal exploration processes enable agents to autonomously sample goals to explore efficiently complex environments with high-dimensional continuous actions. They have been applied successfully to real world robots to discover repertoires of policies producing a wide diversity of effects. Often these algorithms relied on engineered goal spaces but it was recently shown that one can use deep representation learning algorithms to learn an adequate goal space in simple environments. However, in the case of more complex environments containing multiple objects or distractors, an efficient exploration requires that the structure of the goal space reflects the one of the environment. In this paper we show that using a disentangled goal space leads to better exploration performances than an entangled one. We further show that when the representation is disentangled, one can leverage it by sampling goals that maximize learning progress in a modular manner. Finally, we show that the measure of learning progress, used to drive curiosity-driven exploration, can be used simultaneously to discover abstract independently controllable features of the environment. The code used in the experiments is available at \url{https://github.com/flowersteam/Curiosity_Driven_Goal_Exploration}. 

\end{abstract}



\section{Introduction}
A key challenge of lifelong learning is how embodied agents can discover the structure of their environment and learn what outcomes they can produce and control. Within a developmental perspective \citep{Baldassarre2013, Cangelosi2018}, this entails two closely linked challenges. The first challenge is that of exploration: how can learners self-organize their own exploration curriculum to discover efficiently a maximally diverse set of outcomes they can produce. The second challenge is that of learning disentangled representations of the world out of low-level observations (e.g. pixel level), and in particular, discovering abstract high-level features that can be controlled independently.

\textbf{Exploring to discover how to produce diverse sets of outcomes}. Discovering autonomously a diversity of outcomes that can be produced on the environment through rolling out motor programs has been shown to be highly useful for embodied learners. This is key for acquiring world models and repertoires of parameterized skills \citep{Baranes2013, da2014active,hester2017intrinsically}, to efficiently bootstrap exploration for deep reinforcement learning problems with rare or deceptive rewards \citep{conti2017improving, colas18}, or to quickly repair strategies in case of damages \citep{cully2015robots}. However, this problem is particularly difficult in high-dimensional continuous action and state spaces encountered in robotics given the strong constraints on the number of samples that can be experimented. In many cases, naive random exploration of motor commands is highly inefficient due to high-dimensional action spaces, redundancies in the sensorimotor system, or to the presence of ``distractors'' that cannot be controlled \cite{Baranes2013}.

Several approaches to organize exploration can be considered. First, imitation learning can be used to take advantage of observations of another agent acting on the environment \citep{argall2009survey}. While observing the environment changing as a consequence of other agent's actions can often be leveraged, there are many cases where it is either impossible for other agents to demonstrate how to act, or for the learner to observe the motor program used by the other agent. For these reasons, various forms of autonomous curiosity-driven learning approaches have been proposed \citep{oudeyer2018computational}, often inspired by forms of spontaneous exploration displayed by human children \citep{berlyne1966curiosity}. Some of these approaches have used the framework of (deep) reinforcement learning, considering intrinsic rewards valuing states or actions in terms of novelty, information gain, or prediction errors, e.g. \citep{bellemare2016unifying, machado2017laplacian, barto2013intrinsic, hester2017intrinsically}. However, many of these approaches are not directly applicable to high-dimensional redundant continuous action spaces \citep{bellemare2016unifying, pathak2017curiosity}, or face complexity challenges to be applicable to real world robots \citep{houthooft2016curiosity, tang2016exploration}.

Another approach to curiosity-driven exploration is known as Intrinsically Motivated Goal Exploration Processes (IMGEPs) \citep{Baranes2013, Forestier2017}, an architecture closely related to Goal Babbling \citep{Rolf2010}. The general idea of IMGEPs is to equip the agent with a goal space, where each point is a vector of (target) features of behavioural outcomes. During exploration, the agent samples goals in this goal space according to a certain strategy. A powerful strategy for selecting goals is to maximize empirical competence progress using multi-armed bandits \citep{Baranes2013}. This enables to automate the formation of a learning curriculum where goals are progressively explored from simple to more complex, avoiding goals that are either too simple or too complex. For each sampled goal the agent dedicates a budget of experiments to improve its performance regarding this particular goal. IMGEPs are often implemented using a population approach, where the agent stores an archive of all the policy parameters and the corresponding outcomes. This makes the approach powerful since the agent is able to leverage each encountered past experience when facing a new goal. This approach has been shown to enable high-dimensional robots to learn very efficiently locomotion skills \citep{Baranes2013}, manipulation of soft objects \citep{Rolf2010, nguyen2014socially} or tool use \citep{Forestier2017}. Related approaches were recently experimented in the context of Deep Reinforcement Learning, such as in Hindsight Experience Replay \citep{Andrychowicz2017} and Reverse Curriculum Learning \citep{florensa2017reverse} (however using monolithic goal parameterized policies), and within the Power Play framework \citep{schmidhuber2013powerplay}. 

\textbf{Learning disentangled representations of goal spaces.} Even if IMGEP approaches have been shown to be very powerful, one limit has been to rely on engineered representations of goal spaces. For example, experiments in \citep{Baranes2013, colas18, Forestier2017, florensa2017reverse, Andrychowicz2017} have leveraged the availability of goal spaces that directly encoded the position, speed or trajectories of objects/bodies. A major challenge is how to learn goal spaces in cases where only low-level perceptual measures are available to the learner (e.g. pixels). A first step in this direction was presented in \citep{Pere2018}, using deep networks and algorithms such as Variational AutoEncoders (VAEs) to learn goal spaces as a latent representation of the environment. In simple simulated scenes where a robot arm learned to interact with a single controllable object, this approach was shown to be as efficient as using handcrafted goal features. But \citep{Pere2018} did not study what was the impact of the quality of the learned representation. Moreover, when the environment contains several objects including a distractor object, an efficient exploration of the environment is possible only if the structure of the goal space reflects the one of the environment. For example, when objects are represented as abstract distinct entities, modular curiosity-driven goal exploration processes can be leveraged for efficient exploration, by focusing on objects that provide maximal learning progress, and avoiding distractor objects that are either trivial or not controllable \citep{Forestier2017}. An open question is thus whether it is possible to learn goal spaces with adequate disentanglement properties and develop exploration algorithms that can leverage those learned disentangled properties from low-level perceptual measures.

\textbf{Discovering high-level controllable features of the environment.} Although methods to learn disentangled representation of the world exist \citep{Higgins2016, Chen2016a, Hadjeres}, they do not allow to distinguish features that are controllable by the learner from features describing external phenomena that are outside the control of the agent. However, identifying such independantly controllable features \cite{Thomas2017} is of paramount importance for agents to develop compact world models that generalize well, as well as to grow efficiently their repertoire of skills. One idea to address this challenge, initially explored in \citep{Oudeyer2007a}, is that learners may identify and characterize controllable sets of features as sensorimotor space manifolds where it is possible to learn how to control perceptual values with actions, i.e. where learning progress is possible. Unsupervised learning approaches could then build coarse categories distinguishing the body, controllable objects, other animate agents, and uncontrollable objects as entities with different learning progress profiles \citep{Oudeyer2007a}. However, this work only considered identifying learnable and controllable manifolds among sets of \textit{engineered} features. 


In this paper, we explore the idea that a useful learned representation for efficient exploration would be a factorized representation where each latent variable would be sensitive to changes made in a single true dregree of freedom of the environment, while being invariant to changes in other degrees of freedom \citep{Bengio2013}. Further on, we investigate how independently controllable features of the environment can be identified among these disentangled variables through interactions with the environement. We study this question using $\beta$-VAEs \citep{Higgins2016, Higgins2017a} which is a natural extension of VAEs and have been shown to provide good disentanglement properties. We extend the experimental framework of \citep{Pere2018}, simulating a robot arm learning how it can produce outcomes in a scene with two objects, including a distractor. In order to assess the role of the representation we use a two-stage process, which first learns to see and then learns to act. The first stage consists of a representation learning phase where the agent builds a representation of the world by passively observing it (events in the environment are assumed to be produced by another agent in this phase, see \citep{Pere2018}. In the second phase the agent uses this representation to interact with the world, by sampling goals that provide high learning progress, and where goals are target values of one or several latent variables to be reached through action. This procedure was adopted for two reasons. For one, it is similar to the developmental progression in infant development, where the infant first spends most of his time observing the world due to limitations in motor exploration. Secondly, it helps in understanding the impact of disentanglement given the multiple components of the architecture.

\begin{figure}
\includegraphics[width=\textwidth]{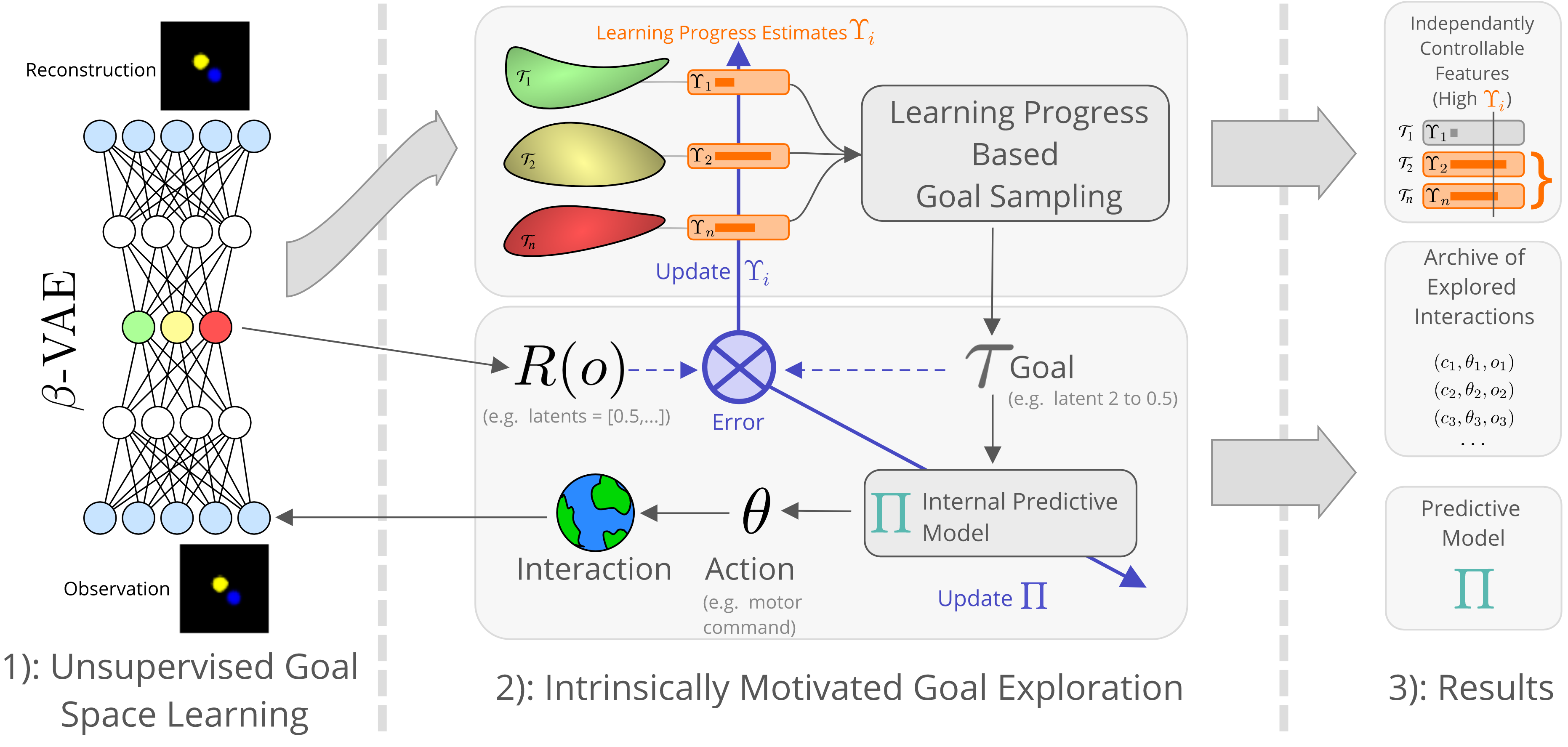}
\caption{The IMGEP-MUGL approach.}
\label{fig:imgep_ugl}
\end{figure}

\textbf{The first contribution} we make in this paper is to study the impact of using a learned disentangled goal space representations on the efficiency of exploration and discovery of a diversity of outcomes in IMGEPs. To the best of our knowledge, it is the first time that the role of disentanglement is studied in the context of exploration. Precisely, we show that:
\begin{compactitem}
\item using a disentangled state representation is beneficial to exploration: using IMGEPS, the agents explores more states in fewer experiments than when the representation is entangled.

\item disentangled representations learned by $\beta$-VAEs can be further leveraged by modular curiosity-driven IMGEPs to explore as efficiently as using handcrafted low-dimensional scene features, in experiments that include both controllable and distractor objects. On the contrary, we show that representations learned by VAEs are not sufficiently structured to enable a similarly efficient exploration.
\end{compactitem}


\textbf{The second contribution} of this article is to show that identifying abstract independently controllable features from low-level perception can emerge from a representation learning pipeline where learning disentangled features from passive observations ($\beta$-VAEs) is followed by curiosity-driven active exploration driven by the maximization of learning progress. This second phase allows in particular to distinguish features related to controllable objects (disentangled features with high learning progress) from features related to distractors (disentangled features with low learning progress).

\section{Modular goal exploration with learned goal spaces}

This section introduces \emph{Intrinsically Motivated Goal Exploration Processes} with modular goal spaces as they are typically used in environments with handcrafted goal spaces. It then describes the architecture used in this article where the handcrafted goal space is replaced by a representation of the space that is learned before exploration and then used as a goal space for IMGEPs. The overall architecture is summarized in \figurename~\ref{fig:imgep_ugl}.

\subsection{Intrinsically motivated goal exploration processes with modular goal spaces}

To fully understand the IMGEP approach, one must imagine the agent as performing a sequence of contextualized and parameterized experiments. The problem of exploration is defined using the following elements:
\begin{compactitem}
\item A context space $\mathcal{C}$. The context $c$ represents the initial state of the environment. It corresponds to parameters of the experiment that are not chosen by the agent. 
\item A parameterization space $\Theta$. The parameterization $\theta$ corresponds to the parameters of the experiment that the agent can control at will (e.g. motor commands for a robot).
\item An observation space $\mathcal{O}$. Here we consider an observation $o$ to be a vector representing all the signals captured by the agent sensors during an experiment (e.g. raw images).
\item An environment dynamic $D: \mathcal{C}, \Theta \rightarrow \mathcal{O}$ which maps parameters performed in a certain context, to observations (or outcomes). This dynamic is considered unknown.
\end{compactitem}

\begin{figure}
  \centering
  \includegraphics[width=.7\textwidth]{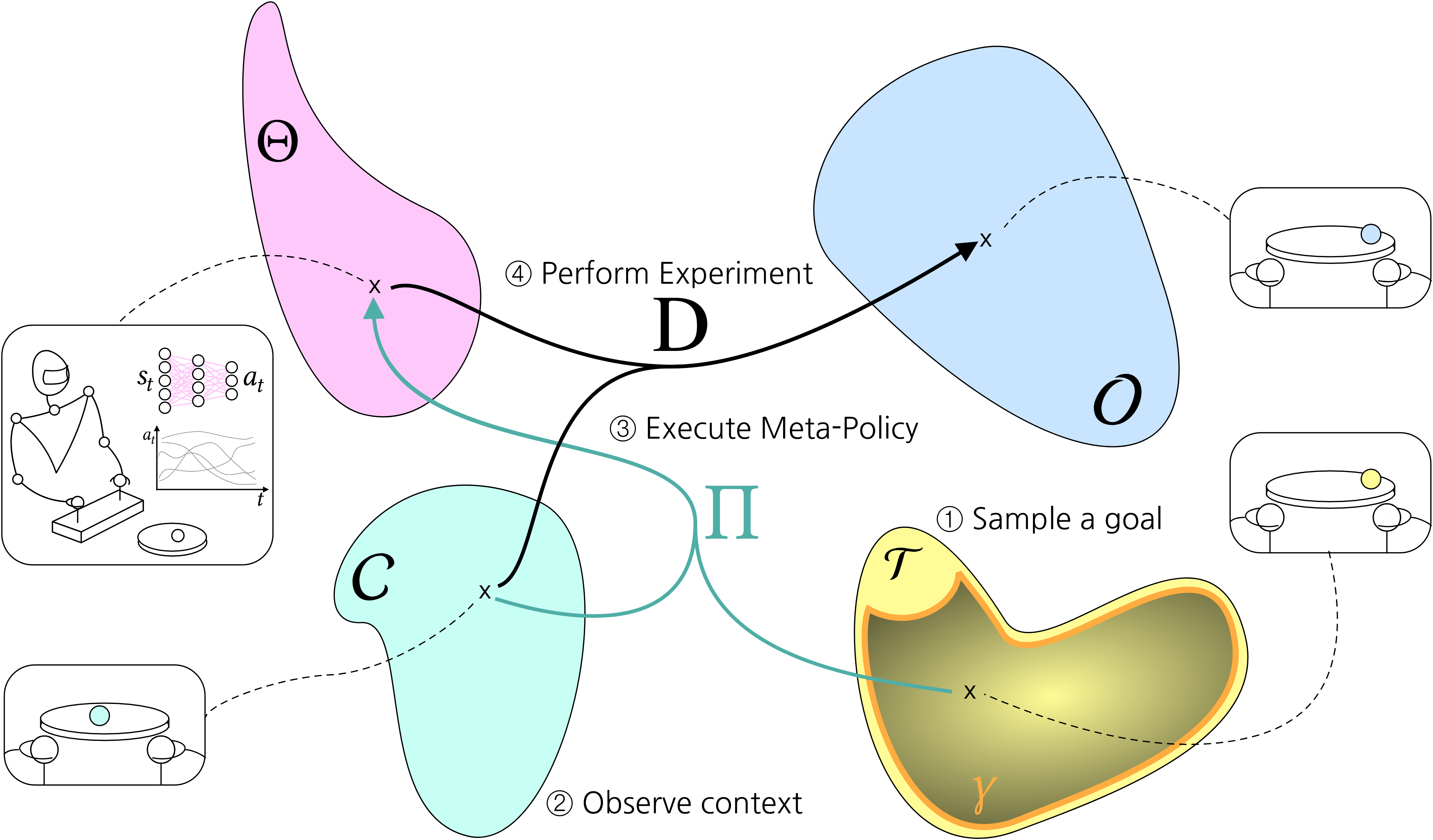}
  \caption{Intrinsically Motivated Goal Exploration Process examplified.}
  \label{fig:imgep}
\end{figure}

For instance, as presented in \figurename~\ref{fig:imgep}, a parameterization could be the weights of a closed-loop neural network controller for a robot manipulating a ball. A context could be the initial position of the ball and an observation could be the position of the ball at the end of a fixed duration experiment. The exploration problem can then be simply put as: 
\begin{quote}
Given a budget of $n$ experiments to perform, how to gather tuples $\{(c_i, \theta_i, o_i)\}_{i=1...n}$ which maximize the diversity of the set of observations $\{o_i\}_{i=1...n}$.
\end{quote}


One approach that was shown to produce good exploration performances is Intrinsically Motivated Goal Exploration Processes (IMGEPs). This algorithmic architecture uses the following elements:
\begin{compactitem}
\item A goal space $\mathcal{T}$. The elements $\tau \in \mathcal{T}$ represent the goals that the agent can set for himself. We also use the term \emph{task} to refer to an element of $\mathcal{T}$.
\item A goal sampling policy $\gamma :\mathcal{T}\mapsto[0,1]$. This distribution allows the agent to choose a goal in the goal space. Depending on the exploration strategy being active or fixed, this distribution can evolve during exploration.
\item A Meta-Policy mechanism (or Internal Predictive Model) $\Pi :\mathcal{T},\mathcal{C}\mapsto\Theta$, which given a goal and a context, outputs a parameterization that is most likely to produce an observation \emph{fulfilling} the goal, under the current knowledge.
\item A cost function $C:\mathcal{T},\mathcal{O}\mapsto\mathbb{R}$, internally used by the Meta-Policy. This cost function outputs the fitness of an observation for a given task $\tau$. 
\end{compactitem}

When the environment is simple, such as for experiments presented in \citep{Pere2018} where a robotic arm explore its possible interactions with a single object, the structure of the goal space is not critical. However, in more complex scenes with multiple objects (e.g. including tools or objects that cannot be controlled), it was shown in \cite{Forestier2016} that it is important to have a goal space which reflects the structure of the environment. In particular, having a modular goal space, i.e. of the form $\mathcal{T} = \bigoplus_{i=1}^N \mathcal{T}_i$, where the $\mathcal{T}_i$ are different modules representing the properties of various objects, leads to much better exploration performances. In that case a goal can correspond to achieving an observation where a given object is in a given position.

The algorithmic architecture works as follows: at each step, the exploration process samples a module, then samples a a goal in this module, observes the context, executes a meta-policy mechanism to guess the best policy parameters for this goal, which it then uses to perform the experiment. The observation is then compared to the goal, and used to update the meta-policy (leveraging the information for other goals) as well as the module sampling policy. The general idea of IMGEPs is depicted in \figurename~\ref{fig:imgep}. Depending on the algorithmic instantiation of this architecture, different Meta-Policy mechanisms can be used \citep{Baranes2013,Forestier2016}. In any case, the Meta-Policy must be initialized using a buffer of experiments $\{c_i, \theta_i, o_i\}$ containing at least two different $o_i$. As such, a bootstrap of several \emph{Random Parameterization Exploration} iterations is always performed at the beginning. This leads to Algorithmic Architecture \ref{alg:imgep}. The reader can refer to Appendix \ref{ann:imgep} for a detailed explanation of the Meta-Policy implementation.

\begin{algorithmicarchitecture}
  \caption{Curiosity Driven Modular Goal Exploration Strategy}
  \label{alg:imgep}
   \KwIn{\\
         Goal modules (engineered or learned with MUGL): $\{R, P_i, \gamma(\cdot| i), C_i\}$, Meta-Policy $\Pi$, History~$\mathcal{H}$}
   \Begin{
       \For{A fixed number of Bootstrapping iterations}{
         Observe context $c$\\
         Sample $\theta \sim \mathcal{U}(-1, 1)$\\
         Perform experiment and retrieve observation $o$\\
         Append $(c, \theta, o)$ to $\mathcal{H}$}
       Initialize Meta-Policy $\Pi$ with history $\mathcal{H}$ \\
       Initialize module sampling probability $p = \mathcal{U}(n_{mod})$ \\
       \For{A fixed number of Exploration iterations}{
         Observe context $c$\\
         Sample a module $i \sim p$\\
         Sample a goal for module $i$, $\tau \sim \gamma(\cdot| i)$ \\
         Compute $\theta$ using Meta-Policy $\Pi$ on tuple $(c, \tau, i)$ \\
         Perform experiment and retrieve observation $o$ \\
         Append $(c, \theta, o)$ to $\mathcal{H}$ \\
         Update Meta-Policy $\Pi$ with $(c, \theta, o)$ \\
         Update module sampling probability $p$ according to Eq.~\eqref{eq:ProbModule}
         }
       }
  \KwRet{The history $\mathcal{H}$}
\end{algorithmicarchitecture}

In a modular architecture the goal sampling policy reads:
\begin{align}
\gamma(\tau) = \gamma(\tau|i) p(i),
\end{align}
where $p(i)$ is the probability to sample the $\mathcal{T}_i$ module, and $\gamma(\tau | i)$ is the probability to sample the goal $\tau$ given that the module $i$ was selected. The strength of the modular architecture is that modules can be selected using a curiosity-driven \emph{active} module sampling scheme. In this scheme, $\gamma(\tau|i)$ is fixed, and $p(i)$ is updated at time $t$ according to:
\begin{align}
\label{eq:ProbModule}
p(i):= 0.9 \times \frac{\Upsilon_i(t)}{\sum_{k=1}^N\Upsilon_k(t)} + 0.1 \times \frac{1}{N},
\end{align} 
where $\Upsilon_i(t)$ is an \emph{interest} measure based on the estimation of the average \textit{improvement} of the precision of the meta-policy for fulfilling goals in $\mathcal{T}_i$, which is a form of learning progress called \textit{competence progress} (see \citep{Baranes2013} and Appendix \ref{ann:imgep} for further details on the interest measure). The second term of Equation \eqref{eq:ProbModule} forces the agent to explore a random module 10\% of the time. The general idea is that monitoring the learning progress allows the agent to concentrate on objects which can be learned to control while ignoring objects that cannot.

\subsection{Modular Unsupervised Goal-space Learning for IMGEP}\label{sec:MGE}

In \cite{Pere2018}, an algorithm for \emph{Unsupervised Goal-space Learning} (UGL) was proposed. The principle is to let the agent observe another agent producing a diversity of observations $\{o_i\}$. This set of observations is used to learn a low-dimensional representation which is then employed as a goal-space. In these experiments, there is always a single goal space corresponding to the learned representation of the environment. However, if one wishes to use the algorithm presented in the previous section, it is necessary to have different goal spaces: one for each module.

\begin{figure}
\begin{subfigure}[t]{.3\textwidth}
\includegraphics[height=17em]{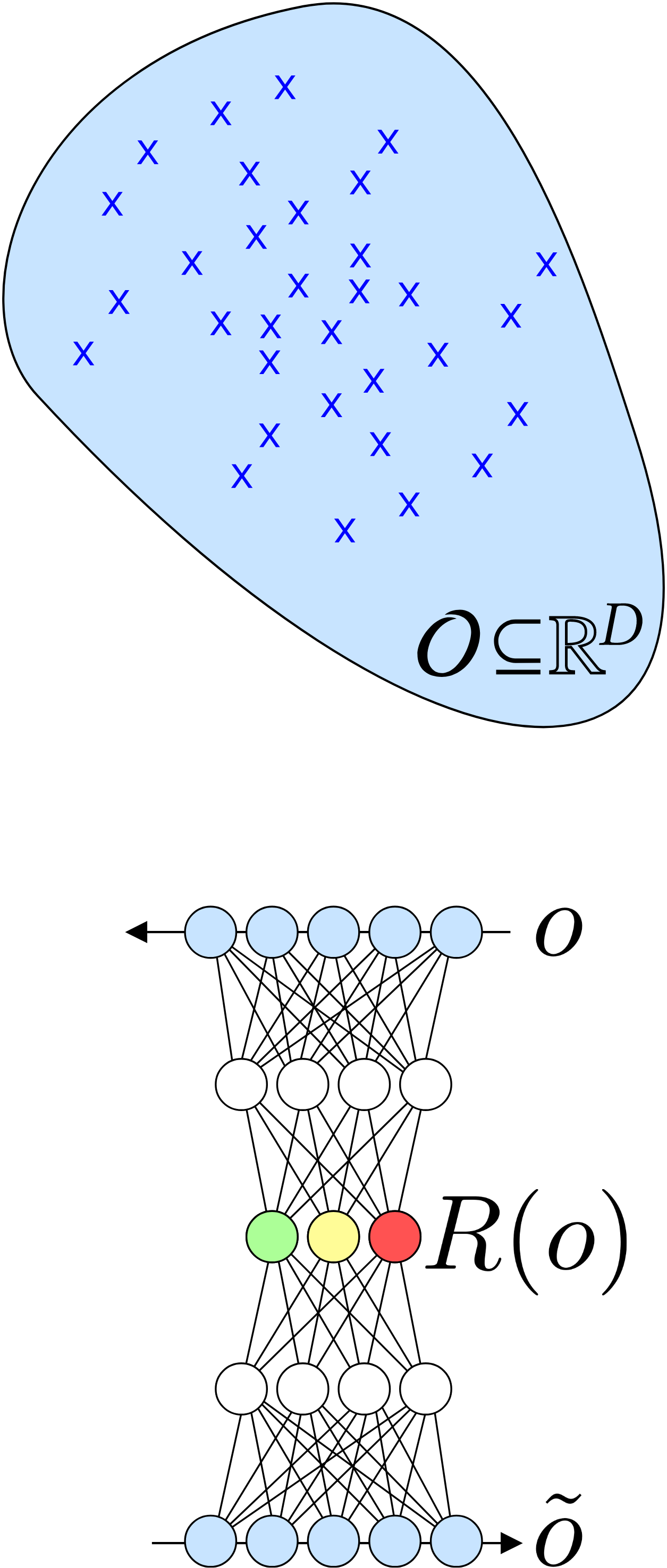}
\caption{Representation learning on observations (Lines 2-5)}
\end{subfigure}
\
\
\begin{subfigure}[t]{.3\textwidth}
\includegraphics[height=17em]{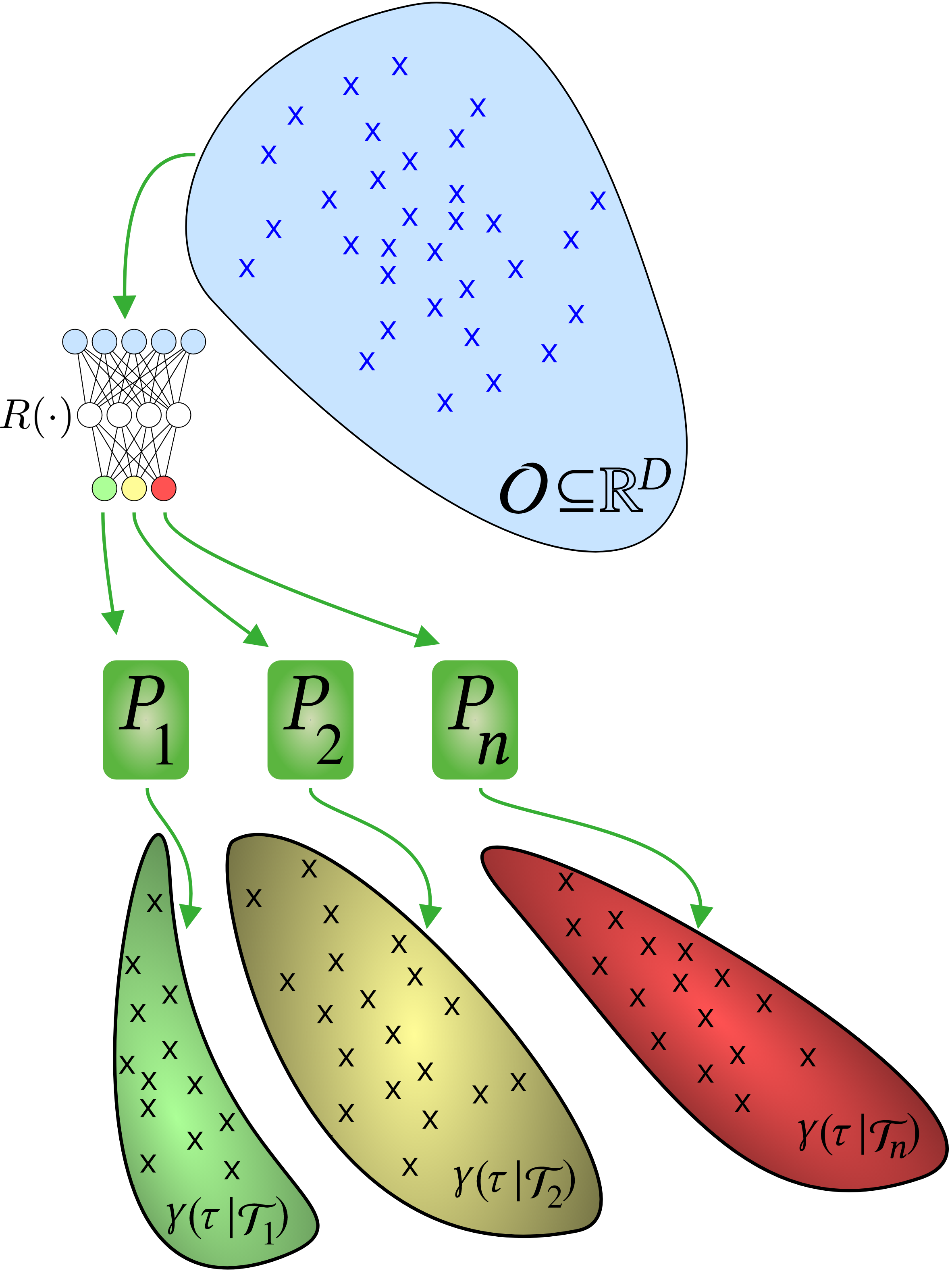}
\caption{Generation of projection operators and estimation of $\gamma(\tau|i)$ distributions (Lines 6-7)}
\end{subfigure}
\
\
\begin{subfigure}[t]{.3\textwidth}
\includegraphics[height=17em]{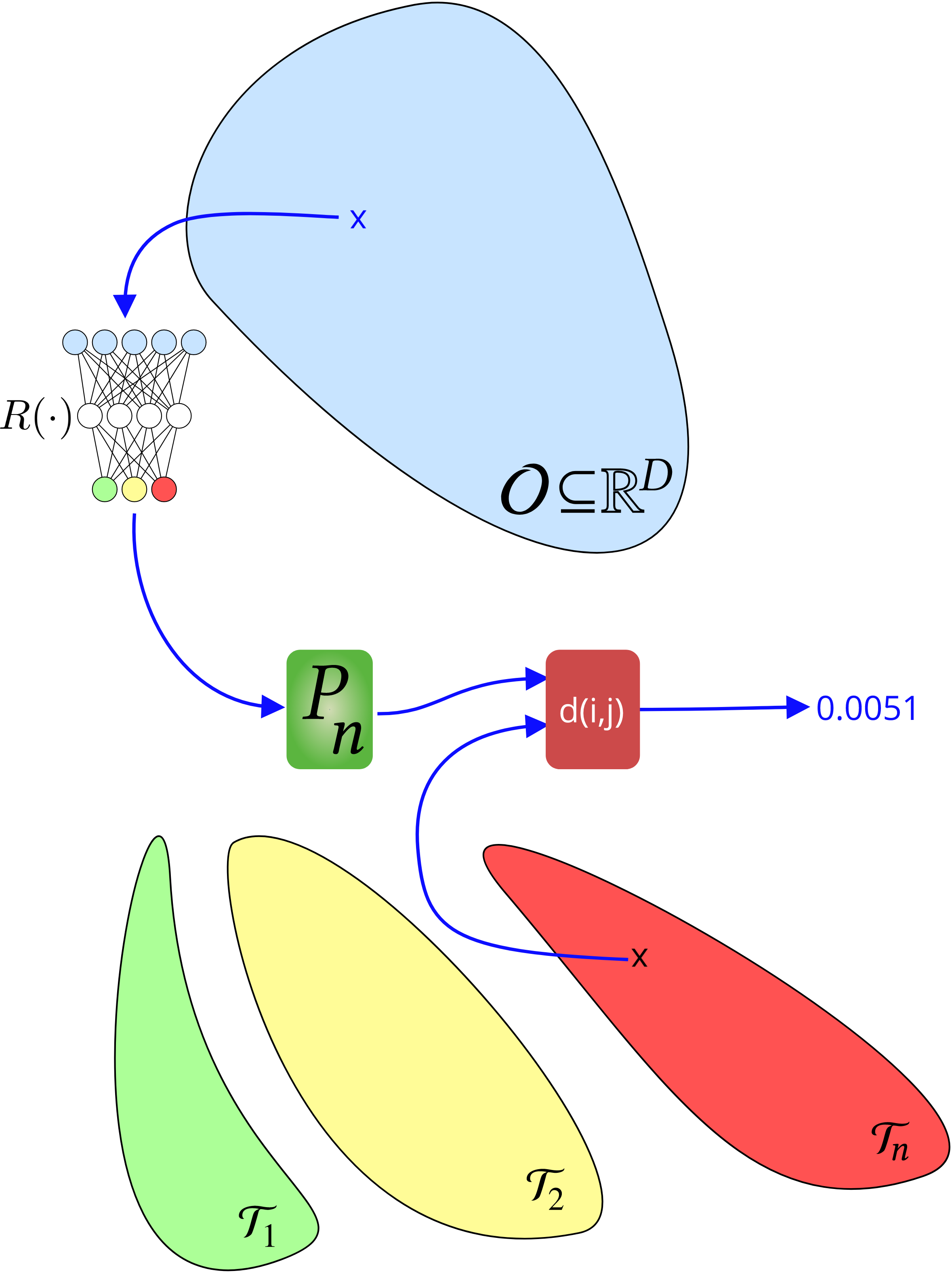}
\caption{Cost function generation (Line~8)}
\end{subfigure}
\caption{The three main steps of the MUGL algorithm}
\label{fig:mugl}
\end{figure}

In order to use a Modular Goal Exploration strategy with a learned goal space, we propose Algorithm \ref{alg:mugl}, which performs \emph{Modular Unsupervised Goal-space Learning} (MUGL) and is represented in \figurename~\ref{fig:mugl}. The idea is to learn a representation of the observations in the same way as UGL. The modules are then defined as subsets of latent variables. For example, a module could be made of the first and second latent variables. Accordingly, goals sampled by this module are defined as achieving a certain value for the first and second latent variables of the representation of an observation. The underlying rationale is that, if we manage to learn a \emph{disentangled} representation of the observations, each latent variable would correspond to a single property of a single object. Thus, by forming modules containing only latent variables corresponding to the same object, the exploration algorithm may be able to explore the different objects separately.

\begin{algorithm}
  \caption{Modular Unsupervised Goal-space Learning (MUGL)}
  \label{alg:mugl}
   \KwIn{\\
       Representation learning algorithm $\mathfrak{R}$ (e.g. VAE, $\beta$VAE), Kernel Density Estimator algorithm $\mathfrak{E}$}
   \Begin{
   	   \For{A fixed number of Observation iterations $n_r$}{
         Observe external agent produce observation $o_i$\\
         Append this sample to database $\mathcal{D}_o = \{o_i\}_{i=0,\dots,n_r}$
       }
     Learn an embedding function $R:\mathcal{O} \mapsto \mathbb{R}^{n_d}$ using algorithm $\mathfrak{R}$ on data $\mathcal{D}_o$\\
     Generate an ensemble of projection operators $\{P_k \}$\\
     Estimate $\gamma(\tau|k)$ from $\{P_k R(o_i)\}_{i=0, \dots,n_r}$ using algorithm $\mathfrak{E}$\\
     Set the cost functions to be $C_k (\tau, o)=\| P_kR(o) - \tau\|$}
  \KwRet{The goal modules $\{R, P_k, \gamma(\tau|k), C_k \}$.}
\end{algorithm}

After learning the representation, a specific criterion is used to decide how the latent variables should be grouped to form modules. In the particular case of VAEs and $\beta$VAEs, the grouping is made by projecting latent variables which have similar Kullback-Leibler on their respective subspace (see Appendix \ref{ann:imgep}). Since representations learned with VAEs and $\beta$VAEs come with a prior over the latent variables, instead of estimating the modular goal-policies $\gamma(\tau|k)$, we used the Gaussian prior assumed during training. Finally, the cost function $C_k(\tau,o)$ is defined, using the distance between the goal and the $k$-th projection of the latent representation of the observation.

The overall approach combining IMGEPs with learned modular goal spaces is summarized in \figurename~\ref{fig:imgep_ugl}. Note that the algorithm proposed in \cite{Pere2018} is a particular instance of this architecture with only one module containing all the latent variables. In this case there is no module sampling strategy, and only a goal sampling strategy. This specific case is here referred to as \emph{Random Goal Exploration} (RGE).

\section{Experiments}

We carried out experiments in a simulated environment to address the following questions:
\begin{compactitem}
\item To what extent is the structure of the learned representation important for the performance of IMGEP-UGL in terms of efficiently discovering a diversity of observations?

\item Is it possible to leverage the structure of the representation with Modular Curiosity-Driven Goal Exploration algorithms?

\item Can the learning progress measure of goal exploration be used to identify controllable abstract features of the environment? 
\end{compactitem}

\paragraph{Environment}

\begin{figure}
\includegraphics[width=\textwidth]{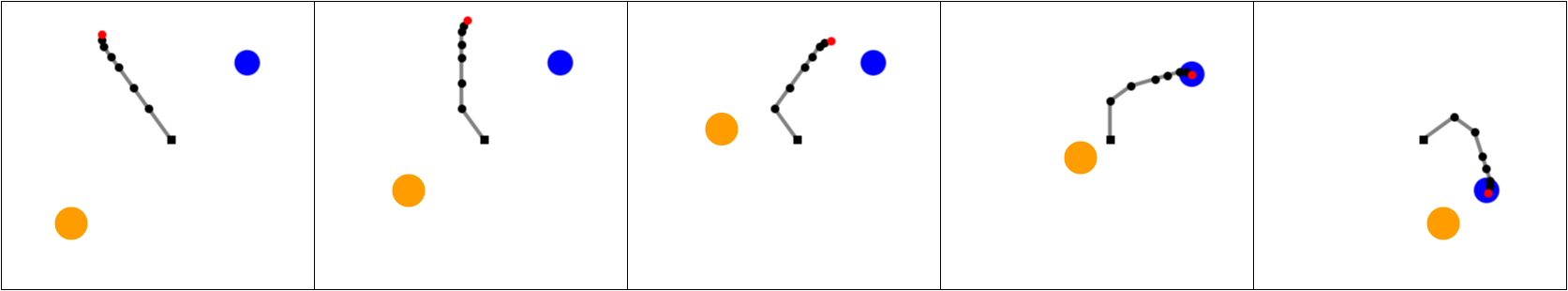}
\caption{A roll-out of experiment in the \textit{Arm-2-Balls} environment. The blue ball can be grasped and moved, while the orange one is a distractor that can not be handled, and follows a random walk.}
\label{fig:armballs}
\end{figure}

We experimented on the \textbf{Arm-2-Balls} environment, where a rotating 7-joints robotic arm evolves in an environment containing two balls of different sizes, as represented in \figurename~\ref{fig:armballs}. One ball can be grasped and moved around in the scene by the robotic arm. The other ball acts as a distractor: it cannot be grasped nor moved by the robotic arm but follows a random walk. The agent perceives the scene as a $64 \times 64$ pixels image. For the representation learning phase, we generated a set of images where the positions of the two balls were uniformly distributed over $[-1, 1]^4$. These images were then used to learn a representation using a VAE or a $\beta$VAE. In order to assess the importance of the disentangled representation, we used the same disentangled/entangled representation for all the instantiations of the exploration algorithms. This allowed us to study the effect of disentangled representations by eliminating the variance due to the inherent difficulty of learning such representations.

\paragraph{Baselines}

Results obtained using IMGEPs with learned goal spaces are compared to two baselines:

\begin{itemize}
\item The first baseline is the naive approach of \textbf{Random Parameter Exploration (RPE)}, where exploration is performed by uniformly sampling parameterizations $\theta$. In the case of hard exploration problems, this strategy is regarded as a low performing one, since no previous information is leveraged to choose the next parameterization. This strategy gives a lower bound on the expected performances of exploration algorithms.

\item The second baseline is \textbf{Modular Goal Exploration with Engineered Features Representation (MGE-EFR)}: it corresponds to a modular IMGEP in which the goal space is handcrafted and corresponds to the true degrees of freedom of the environment. In the \textit{Arm-2-Balls} environment it corresponds to the positions of the two balls, given as a point in $[-1, 1]^4$. Since essentially all the information is available to the agent under a highly semantic form, it is expected to give an upper bound on the performances of the exploration algorithms. We performed experiments with both one module (\textbf{RGE-EFR}) and two modules (one for the ball and one for the distractor) (\textbf{MGE-EFR}).
\end{itemize}

\section{Results}

To assess the performances of the \textbf{MGE} algorithm on learned goal spaces, we experimented with two different representations coming from two learning algorithms: \textbf{$\beta$-VAE} (disentangled) and \textbf{VAE} (entangled). In each case, we ran 20 trials of 10,000 episodes each, for both the \textbf{RGE} and \textbf{MGE} exploration algorithms. One episode is defined as one experimentation/roll-out of a parameter~$\theta$.

\paragraph{Exploration performances}

\begin{figure}
\begin{subfigure}{.5\textwidth}
  \centering
  \includegraphics[width=1.\linewidth]{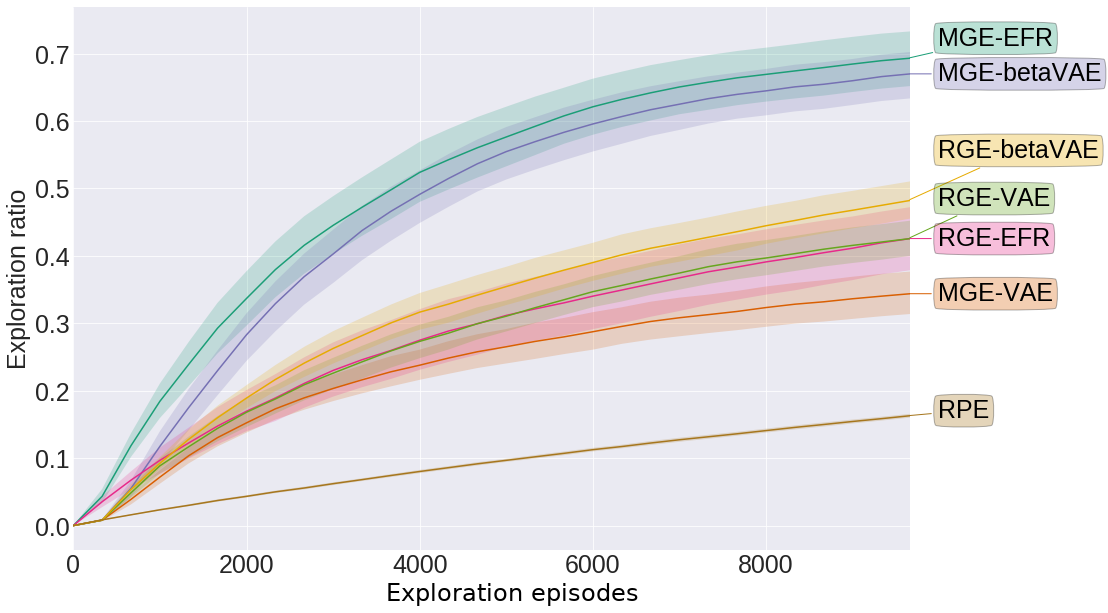}
  \caption{Small exploration noise ($\sigma = 0.05$)}
  \label{fig:ExplorationPerfSmallNoise}
\end{subfigure}
\begin{subfigure}{.5\textwidth}
  \centering
  \includegraphics[width=1.\linewidth]{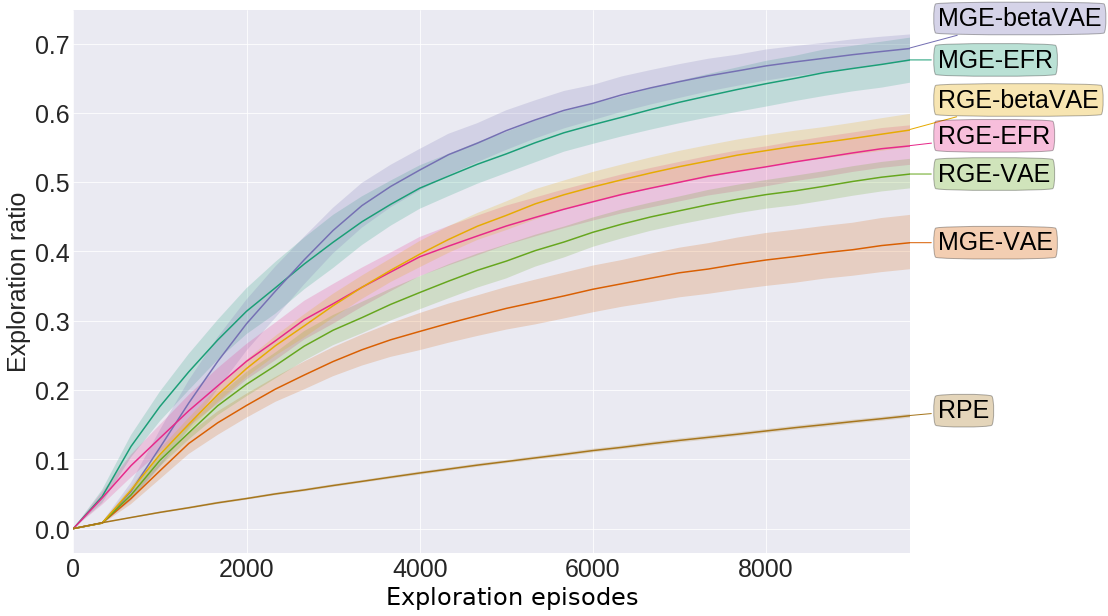}
  \caption{Large exploration noise ($\sigma = 0.1$)}
  \label{fig:ExplorationPerfLargeNoise}
\end{subfigure}%
\caption{Exploration ratio during exploration for different exploration noises.}
\label{fig:ExplorationPerf}
\end{figure}

The exploration performance of all the algorithms was measured according to the number of cells reached by the ball in a discretized grid of 900 cells (30 cells for each dimension of the ball that can be moved; the distractor is not accounted for in the exploration evaluation). Not all cells can be reached given that the arm is rotating and is of unit length: the maximum ratio between the number of reached cells and all the cells is approximately $\pi / 4 \approx 0.8$.

In \figurename~\ref{fig:ExplorationPerf}, we can see the evolution of the ratio of the number of cells visited with respect to all the cells through exploration. First, one can see that all the algorithms have much better performances than the naive \textbf{RPE}, both in term of speed of exploration and final performance. Secondly, for both \textbf{RGE} and \textbf{MGE} with learned goal spaces, using a disentangled representation is beneficial. One can also see that when the representation used as a goal space is disentangled, the modular architecture (\textbf{MGE-$\beta$VAE}) performs much better than the flat architecture (\textbf{RGE-$\beta$VAE}), with performances that match the modular architecture with engineered features (\textbf{MGE-EFR}). However, when the representation is entangled, using a modular architecture is actually detrimental to the performances since each module encodes then only partially for the ball position. \figurename~\ref{fig:ExplorationPerf} also shows that the \textbf{MGE} architectures with a disentangled representation performs particularly well even if the exploration noise is low whereas the \textbf{RGE} architectures or \textbf{MGE} architectures with an entangled representation relies on a large exploration noise to produce a large variety of observations. We cross-refer to Appendix \ref{ann:explo} for examples of exploration curves together with exploration scatters.

\textbf{Benefits of disentanglement and modules}

\begin{figure}
\begin{subfigure}{.5\textwidth}
  \centering
  \includegraphics[width=1.\linewidth]{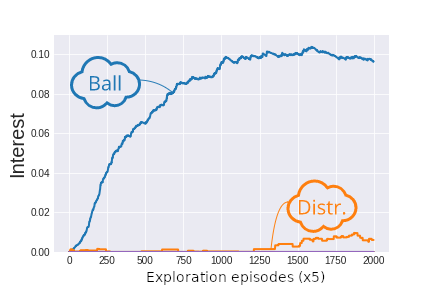}
  \caption{Disentangled representation ($\beta$VAE)}
  \label{fig:InterestBetaVae5Modules}
\end{subfigure}%
\begin{subfigure}{.5\textwidth}
  \centering
  \includegraphics[width=1.\linewidth]{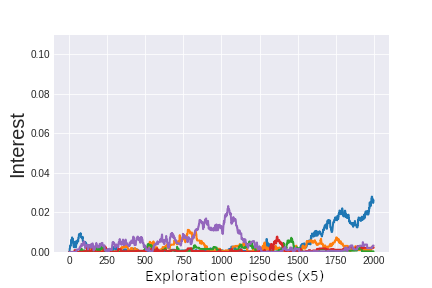}
  \caption{Entangled representation (VAE)}
  \label{fig:InterestVae5Modules}
\end{subfigure}
\caption{Interest evolution for each module through epochs. In the case of a disentangled representation the algorithm shows interest only for the module which correspond to latent variables encoding for the position of the ball (which is unknown by the agent, which does not distinguish between the ball and the distractor).}
\label{fig:InterestMGE5Modules}
\end{figure}

The evolution of the interest of the different modules through exploration is represented in \figurename~\ref{fig:InterestBetaVae5Modules}. First, in the disentangled case, one can see that the interest is high only for the modules corresponding to the latent variables encoding for the ball position.\footnote{The semantic mapping between latent variables and external objects is made by the experimenter.} This is natural since these latent variables are the only ones that can be learned to control with motor commands. In the entangled case, the interest of each module follows a random trajectory, with no module standing out with a particular interest. This effect can be understood as follows: the entanglement introduces spurious correlations between the observations and the tasks in \textit{every} module, which bring the interest measures to follow random fluctuations based on the collected observations. These correlations, in turn, lead the agent to sample more frequently policies that in fact did not have any impact on the observation, making the overall performance worse (see Appendix \ref{app:metapolicy} and  \ref{ann:noisedistractor} for details).

When the representation used as a goal space is disentangled, the modular approach is particularly well suited in the presence of distractors. Indeed, thanks to the projection operator, the noise introduced in the latent variables by the random walk of the distractor is completely ignored by the module which contains the latent variables of the ball. This allows to learn a better inverse model for modules which ignore the distractor, which in turn yields a better exploration (see Appendix \ref{app:metapolicy} and  \ref{ann:noisedistractor} for details).

\textbf{Independently Controllable Features}


As explained above and illustrated in Figure \ref{fig:InterestBetaVae5Modules}, when the representation is disentangled, the MGE algorithm is able to monitor the learnability of certain modules (possibly individual latent features, see \ref{ann:interest}), and leverage it to focus exploration on goals with high learning progress. This is illustrated on the interest curves by the clear difference in interest between modules where learning progress happens and those where it does not. It happens that modules that produce high learning progress correspond precisely to modules that can be controlled. As such, as a side benefit of using modular goal exploration algorithms, the agent discovers in an unsupervised manner which are the features of the environment that can be controlled (and in turn explores them more). This knowledge could then be used by another algorithm whose performance depends on its ability to know which are the independantly controllable features of the environment.



\section{Conclusion}

In this paper we studied the role of the structure of learned goal space representations in IMGEPs. More specifically, we have shown that when the representation possesses good disentanglement properties, they can be leveraged by a curiosity-driven modular goal exploration architecture and lead to highly efficient exploration. In particular, this enables exploration performances as good as when using engineered features. In addition, the monitoring of learning progress enables the agent to discover which latent features can be controlled by its actions, and focus its exploration by setting goals in their corresponding subspace. 

The perspectives of this work are twofold. First it would be interesting to show how the initial representation learning step could be performed online. Secondly, beyond using learning progress to discover controllable features during exploration, it would be interesting to re-use this knowledge to acquire more abstract representations and skills. 


Finally, as mentioned in the introduction, another advantage of using a disentangled representation is that, as was shown in \citep{Higgins2017a}, it evinces superior performances in a transfer learning scenario. Both approaches are not incompatible and one could envision a scheme where one would learn a disentangled representation in a simulated environment and use this representation to perform exploration in a real world environment.



\clearpage
\acknowledgments{We would like to thank Olivier Sigaud for helpful comments on an earlier version of this article.}


\bibliography{papers}  


\clearpage

\section{Appendices}

\subsection{Intrinsically Motivated Goal Exploration Processes}\label{ann:imgep}

In this part, we give further explanations on Intrinsically Motivated Goal Exploration Processes. 

\paragraph{Meta-Policy Mechanism}\label{app:metapolicy}

This mechanism allows, given a context $c$ and a goal $\tau$, to find the parameters $\theta$ that are most likely to produce an observation $o$ \emph{fulfilling} the task $\tau$. The notion of an observation $o$ \emph{fulfilling} a task $\tau$ is quantified using a cost function $C: \mathcal{T} \times\mathcal{O} \mapsto \mathbb{R}$. The cost function can be seen as representing the fitness of the observation $o$ regarding the task $\tau$.




\begin{figure}
\centering
\begin{subfigure}{.7\textwidth}
	\includegraphics[width=\textwidth]{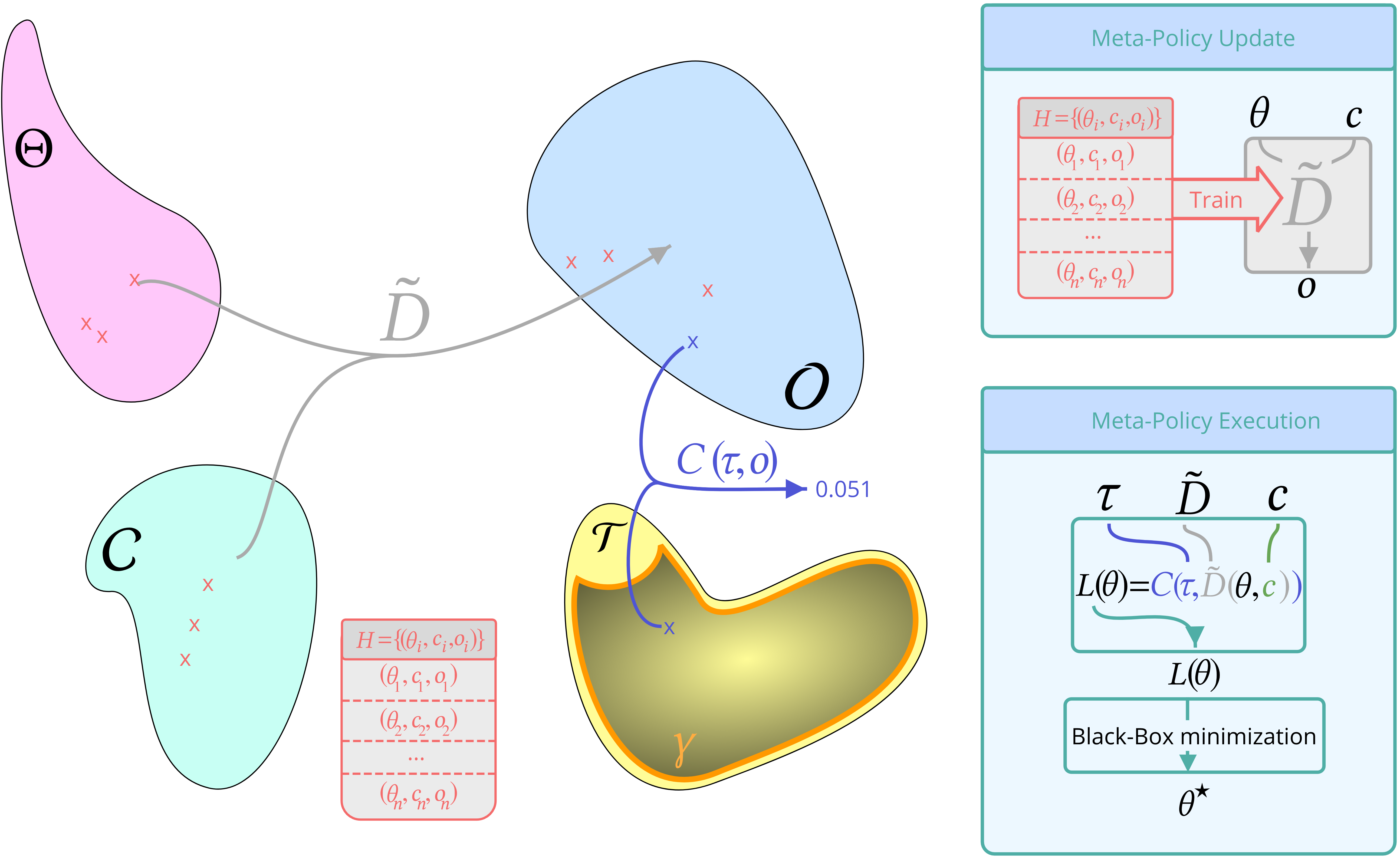}
	\caption{Direct-Model Meta-Policy}
\end{subfigure}
\begin{subfigure}{.7\textwidth}
	\includegraphics[width=\textwidth]{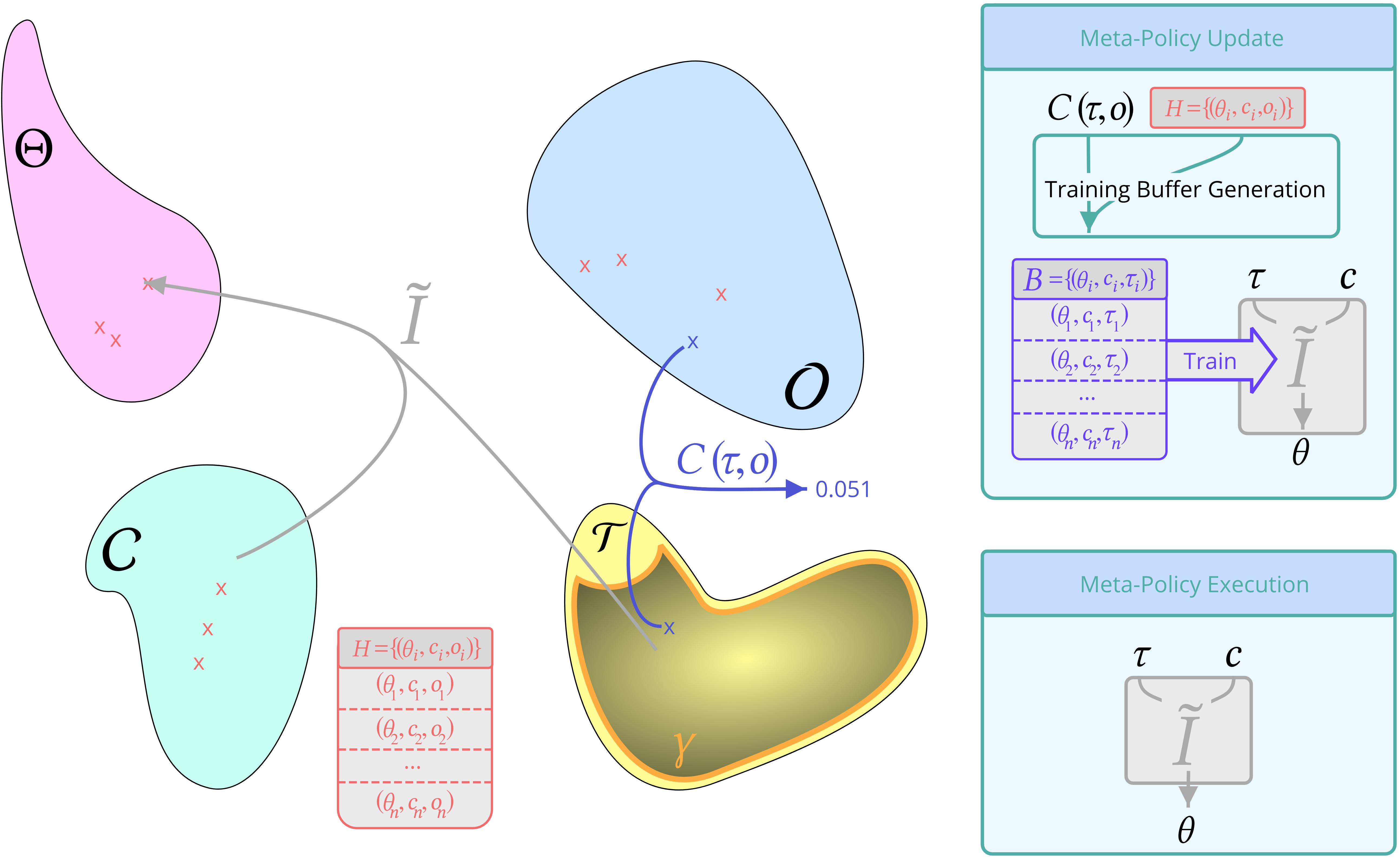}
	\caption{Inverse-Model Meta-Policy}
\end{subfigure}
\caption{The two different approaches to construct a meta-policy mechanism.}
\label{fig:meta-policies}
\end{figure}

The meta-policy can be constructed in two different ways which are depicted in \figurename~\ref{fig:meta-policies}:
\begin{compactitem}
\item \textbf{Direct-Model Meta-Policy:} In this case, an approximate phenomenon dynamic model $\tilde{D}$ is learned using a regressor (e.g. LWR). The model is then updated regularly by performing a training step with the newly acquired data. At execution time, for a given goal $\tau$, a loss function is defined over the parameterization space through $L(\theta)=C(\tau, \tilde{D}(\theta, c))$. A black-box optimization algorithm, such as L-BFGS, is then used to optimize this function and find the optimal set of parameters $\theta$ (see \citep{Baranes2013, Forestier2016, Benureau2016} for examples of such meta-policy implementations in the IMGEP framework). 

\item \textbf{Inverse-Model Meta-Policy:} Here, an inverse model $\tilde{I}: \mathcal{T} \times \mathcal{C} \mapsto \Theta$ is learned from the history $\mathcal{H}$ which contains all the previous experiments in the form of tuples $(c_i, \theta_i, o_i)$. To do so, every experiments observations $o_i$ must be turned into a task $\tau_i$. The inverse model can then be learned using usual regression techniques from the set $\{( \tau_i, c_i, \theta_i )\}$.
\end{compactitem}


In our case, we took the approach of using an Inverse-Model based Meta-Policy. We draw the attention of the reader on the following implementation details:
\begin{compactitem}
\item Depending on the case, multiple observations, and consequently multiple parameters can optimally solve a task, while a combination of them cannot. This is known as the redundancy problem in robotics and special approaches must be used to handle it when learning inverse models, in particular within the IMGEP framework \cite{Baranes2013}. This has also been tackled under the terminology of multi-modality in \cite{PathakICLR2018}. To solve this problem, we used a $\kappa$-nn regressor with $\kappa=1$.

\item Turning observations $\{o_i\}$ into goals $\{\tau_i\}$ may prove difficult in some cases. Indeed, it may happen that a given observation does not solve optimally any task in the goal space, or that it solves optimally multiple tasks. In our case, we assumed that the learned encoder is a one-to-one map from observation space to goal space and thus, that every observation solves optimally a unique task in each module. Hence, tasks were associated to observations using the encoder $R$: $\tau_i := R(o_i)$.

\item Since the different modules are associated to projection operators, each produced observation $o$ optimally solve one task for each module. Indeed, if we consider projections on the canonical axis of the latent space, $o$ will solve one task for each module, corresponding to each component of $R(o)$. This mechanism allows to leverage information of every single observation, for all goal-space modules. For this reason, one $\kappa$-nearest-neighbor model was used for each module of the goal space. At each exploration iteration all the modules are updated using their associated projection operators on the embedding of the outcome.
\end{compactitem}

Our particular implementation of the Meta-Policy is outlined in Algorithm~\ref{alg:train_meta}. The Meta-Policy is instantiated with one database per goal module. Each database store the representations of the observations projected on its associated subspace together with the associated contexts and parameterizations. Given that the meta policy is implemented with a nearest neighbor regressor, training the meta policy simply amounts to updating all the databases. Note that, as stated above, even though at each step the goal is sampled in only one module, the observation obtained after an exploration iteration is used to update all databases.

\begin{algorithm}[h]
  \caption{Meta-Policy (simple implementation using a nearest-neighbor model)}
  \label{alg:train_meta}
   \textbf{Require:} Goal modules: $\{R, P_k, \gamma(\tau|k), C_k \}_{k \in \{1, .., n_{mod} \} }$ \\
          \SetKwFunction{FMain}{Initialize\_Meta-Policy}
       \SetKwProg{Fn}{Function}{:}{}
       \Fn{\FMain{$\mathcal{H}$}}{
        \For{$k \in \{1, .., n_{mod} \}$}{
        $\text{database}_k \leftarrow \text{VoidDatabase}$ \\
            \For{$(c, \theta, o) \in \mathcal{H}$}{
               Add $(c, \theta, P_kR(o))$ to $\text{database}_k$
          }
      }
}
       \SetKwFunction{FMain}{Update\_Meta-Policy}
       \SetKwProg{Fn}{Function}{:}{}
       \Fn{\FMain{$c, \theta, o$}}{
        \For{$k \in \{1, .., n_{mod} \}$}{
               Add $(c, \theta, P_kR(o))$ to $\text{database}_k$
          }
}
\SetKwFunction{FMain}{Infer\_parameterization}
       \SetKwProg{Fn}{Function}{:}{}
       \Fn{\FMain{$c, \tau, k$}}{
   	       $\theta \leftarrow$ NearestNeighbor$(\text{database}_k, c, \tau)$ \\
           
          \KwRet{$\theta$}
}
\end{algorithm}


\paragraph{Active module sampling based on Interest measure}
Recalling from the paper, at each iteration, the probability of sampling a specific module $\mathcal{T}_i$ is given by:
\begin{align*}
\gamma(i):= 0.9\times\frac{\Upsilon_i(t)}{\sum_{k=1}^N\Upsilon_k(t)} + 0.1\times\frac{1}{N},
\end{align*}
where $\Upsilon_i(t)$ represents the interest of the $\mathcal{T}_i$ module after $t$ iterations. Let $\mathcal{H}^{(i)}_t = \{ (\tau_k, \theta_k, P_iR(o_k)) \}_{\tau_k \in \mathcal{T}_i}$ be the history of experiments obtained when the goal was sampled in module $\mathcal{T}_i$. The progress in module $i$ at exploration step $t$ is defined as:
\begin{align}
	\delta_t^{(i)} = C_i(\tau_t, P_iR(o')) - C_i(\tau_t, P_iR(o_t)),
\end{align}
where $o_t$ and $\tau_t$ are respectively the observation and goal for the current exploration step and $o'$ is the observation associated to the experiment in $\mathcal{H}^{(i)}_t$ for which the goal $\tau'$ is the closest to $\tau_t$. The interest of a module is designed to track the progress. Specifically, the interest of each module is updated according to:
\begin{align}
\Upsilon_i(t) = \frac{n - 1}{n} \Upsilon_i(t - 1) + \frac{1}{n} \delta_t,
\end{align}
where $n = 1000$ is a decay rate that ensures that if no progress is made the interest of the module will go to zero over time. One can refer to \cite{Forestier2016} for details on this approach.

\begin{figure}
    \centering
    \includegraphics[width=0.7\textwidth]{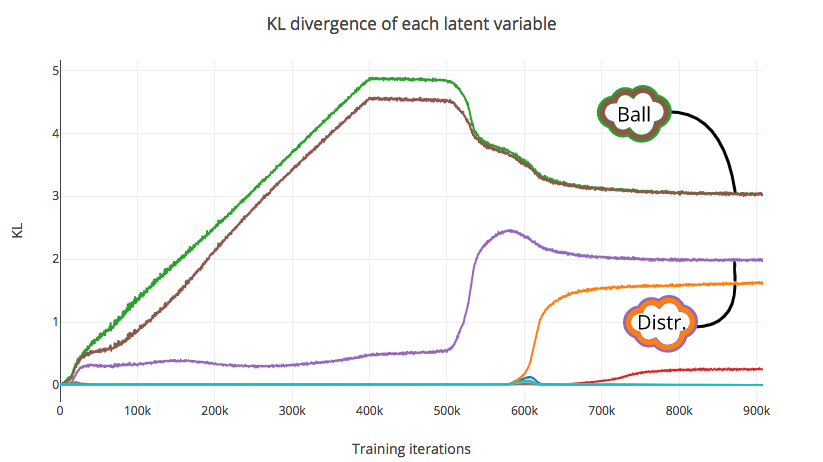}
    \caption{Kullback-Leibler divergence of each latent variable over training.}
    \label{fig:KLlatents}
\end{figure}

\paragraph{Projection criterion for VAE and $\beta$VAE}
An important aspect of the MUGL algorithm is the choice of the projection operators $\{P_k\}$. In this work, the representation learning algorithms are VAE and $\beta$VAE. In this case, two projection schemes can be considered:
\begin{compactitem}
\item \textbf{Projection on all canonical axis:} $n_d$ projection operators, each projecting the latent point on a single latent axis. 
\item \textbf{Projection on 2D planes sorted by $\mathbb{D}_{KL}$:} $\frac{n_d}{2}$ projection operators, each projecting on a 2D plane aligned with latent axis. The grouping of dimensions as 2D planes is performed by sorting the dimensions by increasing $\mathbb{D}_{KL}$, i.e. the divergence is computed for each dimension, by projecting the latent representation on the dimension and measuring its divergence with the unit gaussian prior. Latent dimensions are then grouped two by two according to their $\mathbb{D}_{KL}$ value.
\end{compactitem}
In this work we mainly considered the second grouping scheme. The first grouping scheme could be considered to discover which features can be controlled. Of course in practice one often does not know in advance how many latent variables should be grouped together and it can be necessary to consider more advanced grouping schemes. In practice it is often the case that latent variables which correspond to the same objects have similar KL divergence value (see \figurename~\ref{fig:KLlatents} for an example of a training curve and appendix \ref{ann:drl} for an explanation of this phenomenon). As such it could be envisioned to group latent variables which have similar KL divergence together.

\subsection{Deep Representation Learning Algorithms}\label{ann:drl}

In this section we summarize the theoretical arguments behind Variational AutoEncoder (VAE) and $\beta$VAE.

\paragraph{Variational Auto-Encoders (VAEs)} Let $\mathbf{x} \in \mathcal{X}$ be a set of observations. If we assume that the observed data are realizations of a random variable, we can hypothesize that they are conditioned by a random vector of independent factors $\mathbf{z}$, i.e. that $p(\mathbf{x} , \mathbf{z})= p(\mathbf{z}) p_{\theta}(\mathbf{x}, \mathbf{z})$, where $p(\mathbf{z})$ is a \textit{prior} distribution over $z$ and $p_{\theta}(\mathbf{x}, \mathbf{z})$ is a \textit{conditional distribution}. In this setting, given a i.i.d dataset $X = \{\mathbf{x}^1, \ldots, \mathbf{x}^N \}$, learning the model amount to searching the parameters $\theta$ that maximizes the dataset likelihood:
\begin{equation}
\log\mathfrak{L}(\mathcal{D}) = \sum_{i=1}^N \log p_{\theta} (\mathbf{x}^i)
\end{equation}
However, in most cases, the marginal probability:
\begin{equation}
p_{\theta} (\mathbf{x}) = \int p(\mathbf{x}, \mathbf{z}) \mathrm{d}\mathbf{z}
\end{equation}
and the posterior probability:
\begin{equation}
p_{\theta}(\mathbf{z} \vert \mathbf{x}) = \frac{p(\mathbf{x}, \mathbf{z})}{p(\mathbf{z})} = \frac{p(\mathbf{x}, \mathbf{z})}{\int p(\mathbf{x}, \mathbf{z}) \mathrm{d}\mathbf{z}}
\end{equation}
are both computationally intractable, making the maximum likelihood estimation unfeasible. To overcome this problem, we can introduce an arbitrary distribution $q_{\phi}(\mathbf{z} | \mathbf{x})$ and remark that the following holds:
\begin{equation}
    \log p_{\theta} (\mathbf{x}) = \mathcal{L}(\mathbf{x}; \theta, \phi) + \mathbb{D}_{KL}\left[ q_{\phi}(\mathbf{z}|\mathbf{x})\| p_{\theta}(\mathbf{z}|\mathbf{x}) \right],
    \label{eq:variational_inference}
\end{equation}
where $\mathbb{D}_{KL}$ denotes the Kullback-Leibler (KL) divergence and
\begin{equation}
\mathcal{L}(\mathbf{x}; \theta, \phi) = \mathbb{E}_{\mathbf{z}\sim q_{\phi}(\mathbf{z}|\mathbf{x})}[\log p_{\theta}(\mathbf{x}|\mathbf{z})] - \mathbb{D}_{KL}[q_{\phi}(\mathbf{z}|\mathbf{x}) \| p(\mathbf{z})].
    \label{eq:elbo}
\end{equation}
Since the KL divergence is non-negative, it follows from \eqref{eq:variational_inference} that:
\begin{equation}
\mathcal{L}(\mathbf{x}; \theta, \phi) \leq \log p_{\theta}(\mathbf{x}) - \mathbb{D}_{KL}[ q_{\phi} (\mathbf{z}|\mathbf{x}) \| p_{\theta} (\mathbf{z} | \mathbf{x})]
\end{equation}
for any distribution $q$, hence the name of Evidence Lower Bound (ELBO). Consequently, maximizing the ELBO has the effect to maximize the log likelihood, while minimizing the KL-Divergence between the approximate $q_{\phi}(\mathbf{z}|\mathbf{x})$ distribution, and the true unknown posterior $p_{\theta} (\mathbf{z}|\mathbf{x})$. The approach taken by VAEs is to \emph{learn} the parameters of both conditional distributions $p_{\theta}(\mathbf{x}|\mathbf{z})$ and $q_{\phi}(\mathbf{z}|\mathbf{x})$ as non-linear functions. This is done by maximizing the ELBO of the dataset:
\begin{equation}
\mathcal{L}(\theta, \phi) = \sum_{i=1}^N \mathcal{L} (\mathbf{x}^i; \theta, \phi)
\end{equation}
by jointly optimizing over the parameters $\theta$ and $\phi$. When the prior $p(\mathbf{z})$ is an isotropic unit Gaussian distribution and the variational approximation $q_{\phi}(\mathbf{z} | \mathbf{x})$ follow a Multivariate Gaussian distribution with diagonal covariance, the KL divergence term can be computed in a closed form.

\paragraph{$\beta$ Variational Auto-Encoders ($\beta$VAEs)}

In essence, a VAE can be understood as an AutoEncoder with stochastic units ($q_{\phi}(\mathbf{z} \vert \mathbf{x})$ plays the role of an encoder while $p_{\theta}(\mathbf{x} \vert \mathbf{z})$ plays the role of the decoder), together with a regularization term given by the KL divergence between the approximation of the posterior and the prior. This term ensures that the latent space is structured. The existence of a prior over the latent variables gives the ability to use a VAE as a generative model, and latent variables sampled according to the prior $p(\mathbf{z})$ can be transformed by the decoder into samples.

Ideally, in order to be more easily interpretable, we would like to have a disentangled representation, i.e. a representation where a single latent is sensitive to changes in only one generative factor while being invariant to changes in other factors. When the prior distribution $p(\mathbf{z})$ is an isotropic unit Gaussian distribution ($p(\mathbf{z}) = \mathcal{N}(0, I)$) the role of the regularization term can be understood as a pressure that encourages the VAE to learn independent latent factors $\mathbf{z}$. As such, it was suggested in \cite{Higgins2016, Higgins2017a} that modifying the training objective to:
\begin{equation}
\mathcal{L}(\mathbf{x}; \theta, \phi) = \mathbb{E}_{\mathbf{z}\sim q_{\phi}(\mathbf{z}|\mathbf{x})}[\log p_{\theta}(\mathbf{x}|\mathbf{z})] - \beta \mathbb{D}_{KL}[q_{\phi}(\mathbf{z}|\mathbf{x}) \| p_{\theta}(\mathbf{z})],
    \label{eq:elboBetaVAE}
\end{equation}
where $\beta$ is an additional parameter, will allow one to control the degree of applied pressure to learn independent generating factors by tuning the parameter $\beta$. In particular values of $\beta$ higher than $1$ should lead to representations with better disentanglement properties.


One of the drawbacks of $\beta$VAE is that for large values of $\beta$ the reconstruction cost is often dominated by the KL divergence term. This leads to poor reconstructed samples where the model ignores some of the factors of variation altogether. In order to tackle this issue, it was further suggested in \cite{Burgess2017a} to modify the training objective to be:
\begin{equation}
\mathcal{L}(\mathbf{x}; \theta, \phi) = \mathbb{E}_{\mathbf{z}\sim q_{\phi}(\mathbf{z}|\mathbf{x})}[\log p_{\theta}(\mathbf{x}|\mathbf{z})] - \beta \vert \mathbb{D}_{KL}[q_{\phi}(\mathbf{z}|\mathbf{x}) \| p_{\theta}(\mathbf{z})] - C \vert,
    \label{eq:elboBetaVAECapacity}
\end{equation}
where $C$ is a new parameter that defines the capacity of the VAE. The value of $C$ determines the capacity of the network to encode information in the latent variables. For low values of the capacity the network will mostly reconstruct properties which have a high reconstruction cost whereas high capacity ensures that the network can have a low reconstruction error. By optimizing the training objective \eqref{eq:elboBetaVAECapacity} with a gradually increased capacity the network will start to encode features with high reconstruction cost and then progressively encode more factors of variations whilst retaining disentangling in previously learned factors. At the end of the training one should thus obtain a representation with good disentanglement properties where each factor of variation is encoded into a unique latent variable.

In our experiments we used the training objective of Eq.~\eqref{eq:elboBetaVAECapacity} as detailed in Sec.~\ref{ann:detailsNN}.


\subsection{Disentanglement properties}\label{ann:DisentProp}

\begin{figure}
\begin{subfigure}{.5\textwidth}
  \centering
  \includegraphics[width=.9\linewidth]{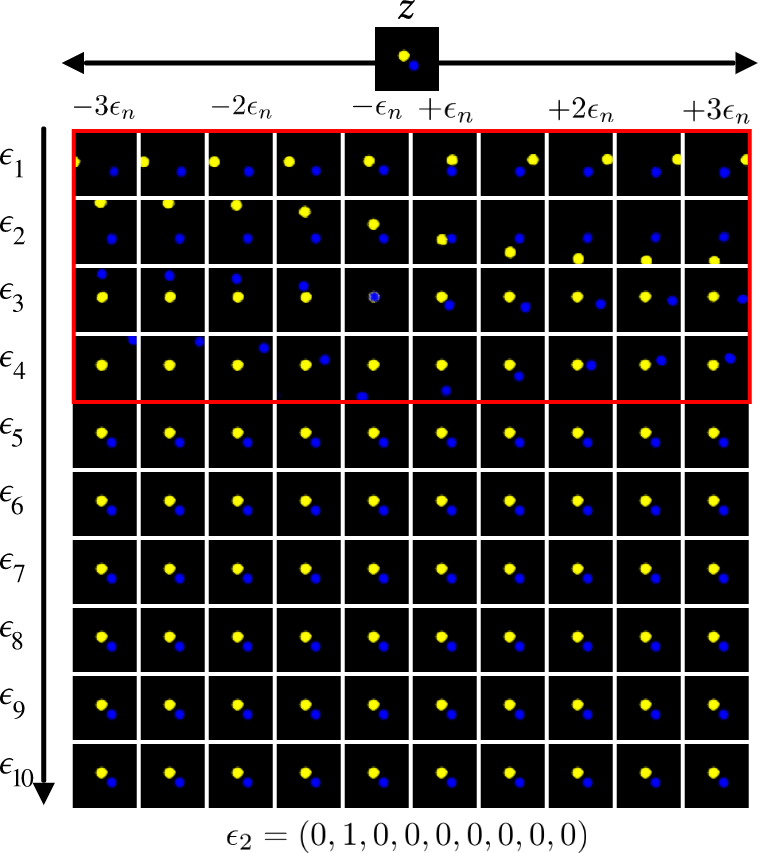}
  \caption{Disentangled latent representation learned by \textbf{$\beta$VAE}}
  \label{fig:LatentTraversalDisent}
\end{subfigure}%
\begin{subfigure}{.5\textwidth}
  \centering
  \includegraphics[width=.9\linewidth]{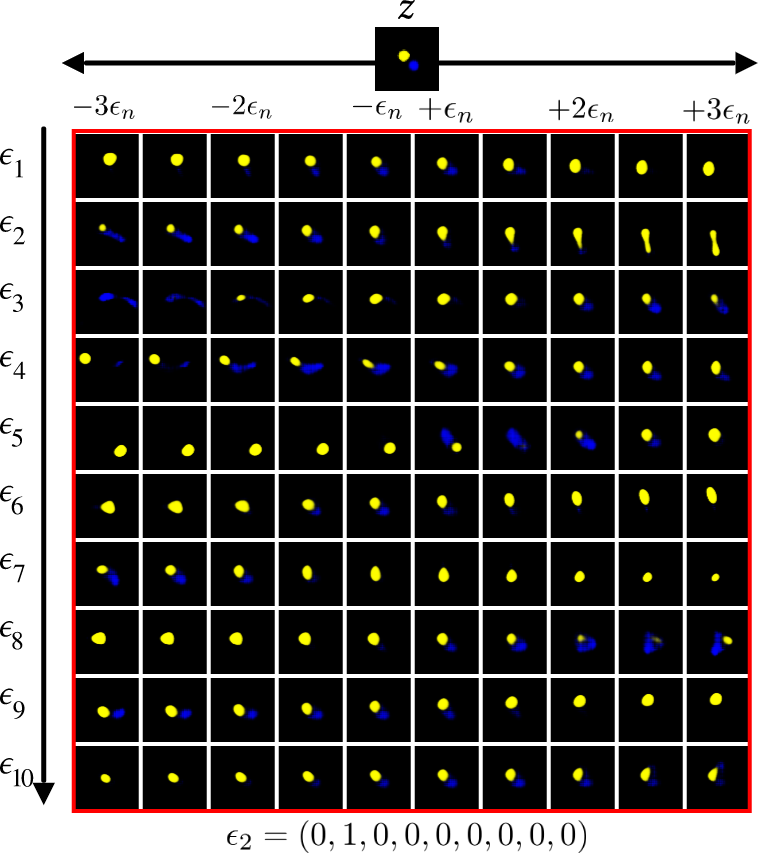}
  \caption{Entangled latent representation learned with \textbf{VAE}}
  \label{fig:LatentTraversalEnt}
\end{subfigure}
\caption{(a) Latent traversal study for a disentangled representation ($\beta$VAE). Each row represents a latent variable and rows are ordered by KL divergence (lowest at the bottom). Each row represents the reconstruction obtained from the traversal of each latent variable over three standard variation around the unit Gaussian prior mean while keeping the other latent variables to the value obtained by running inference on an image of the dataset. From the picture it is clear that the first two latent variables encode the $x$ and $y$ position of the Ball and that the third and fourth latent variables encode the $x$ and $y$ position of the Distractor. At the end of the training the remaining latent variables have converged to the unit Gaussian prior. (b) Similar analysis for an entangled representation (VAE). No latent variable encode for a single factor of variation.}
\label{fig:LatentTraversal}
\end{figure}

We compared the disentanglement properties of two representations. One with the procedure outlined in Sec.~\ref{ann:drl} with $\beta = 150$ and a capacity linearly increased to 12 over the course of the training. The other representation was a vanilla VAE with $\beta = 1$. In order to assess the disentanglement properties of the two representations we performed a latent traversal study. The results of which are displayed in Figure~\ref{fig:LatentTraversal}.

It was experimentally observed that the positions of the two balls were indeed disentangled in most cases when the representation was obtained using a $\beta$VAE even though the data used for the training was generated using independent samples for the position of the two balls. As explained in the previous section, this effect can be understood as follows: since the two balls do not have the same reconstruction cost, the VAE tends to reconstruct the object with the highest reconstruction cost first (in this case the largest ball), and when the capacity reaches the adequate value, it starts reconstructing the other ball~\cite{Burgess2017a}. It follows that the latent variables encoding for the position of the two balls are often disentangled.

\subsection{Details of Neural Architectures and training}\label{ann:detailsNN}

\paragraph{Model Architecture} The encoder for the VAEs consisted of 4 convolutional layers, each with 32 channels, 4x4 kernels, and a stride of 2. This was followed by 2 fully connected layers, each of 256 units. The latent distribution consisted of one fully connected layer of 20 units parametrizing the mean and log standard deviation of 10 Gaussian random variables. The decoder architecture was the transpose of the encoder, with the output parametrizing Bernoulli distributions over the pixels. ReLu were used as activation functions. This architecture is based on the one proposed in \cite{Higgins2016}.


\paragraph{Training details} For the training of the disentangled representation we followed the procedure outlined in Sec.~\ref{ann:drl}. The value of $\beta$ was 150 and the capacity was linearly increased from 0 to 12 over the course of 400,000 training iterations. The optimizer used was Adam \cite{Kingma2015} with a learning rate of $5e^{-5}$ and batch size of 64. The overall training of the representation took 1M training iterations.
For the training of the entangled representation the same procedure was followed except that $\beta$ was set to 1 and that the capacity was set to 0.

\subsection{Interest curves for Projection on all canonical axis}\label{ann:interest}

In the main text of the paper we discussed the case of 5 modules. In general one can imagine having one modules per latent variable. In this case the agent would learn to discover and control each of the latent variables separately.

\begin{figure}
\begin{subfigure}{.5\textwidth}
\includegraphics[width=\textwidth]{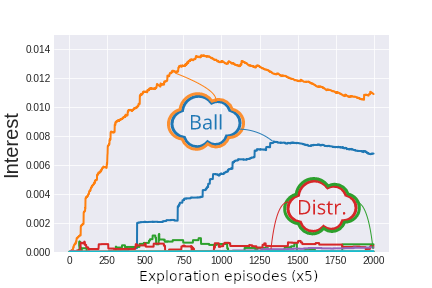}
\caption{Disentangled representation ($\beta$VAE)}
\end{subfigure}
\begin{subfigure}{.5\textwidth}
\includegraphics[width=\textwidth]{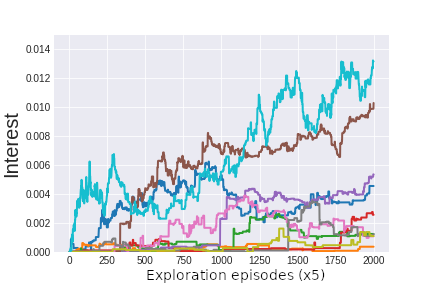}
\caption{Entangled representation (VAE)}
\end{subfigure}
\caption{Interest curves for Projection on all canonical axis}
\label{fig:interest10}
\end{figure}

In \figurename~\ref{fig:interest10} is represented the interest curves when there are 10 modules, one for each latent variable. When the representation is disentangled ($\beta$VAE), the interest is high only for modules which encode for some degrees of freedom of the ball. On the other hand, when the representation is entangled, the interest follows some kind of random walk for all modules. This is due to the fact that all the modules encode for both the ball and the distractor position which introduces some noise in the prediction of each module.

\subsection{Effect of noise in the distractor}\label{ann:noisedistractor}

\begin{figure}
\centering
\includegraphics[width=.7\textwidth]{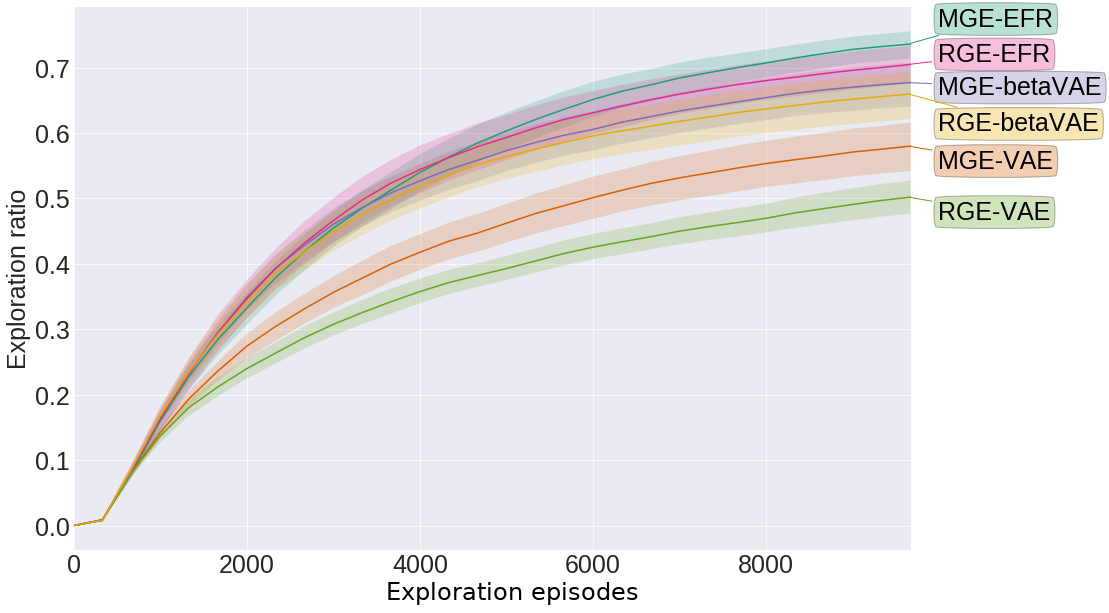}
\caption{Exploration ratio through epochs for all the exploration algorithms in the \textit{Arm-2-Balls} environment with a distractor that does not move.}
\label{fig:ExplorationPerfSmallDistractNoise}
\end{figure}

We also experimented with different noise level in the displacement of the distractor. As expected, when the noise level is low, the distractor does not move very far from its initial position and no longer acts as a distractor. In this case there is no advantage of using a modular algorithm as illustrated by \figurename~\ref{fig:ExplorationPerfSmallDistractNoise}. However, it is still beneficial to have a disentangled representation since it helps in learning good inverse models.

\subsection{Exploration Curves}\label{ann:explo}

\begin{figure}
	\centering
    \begin{subfigure}{1.\textwidth}
    \includegraphics[width=\textwidth]{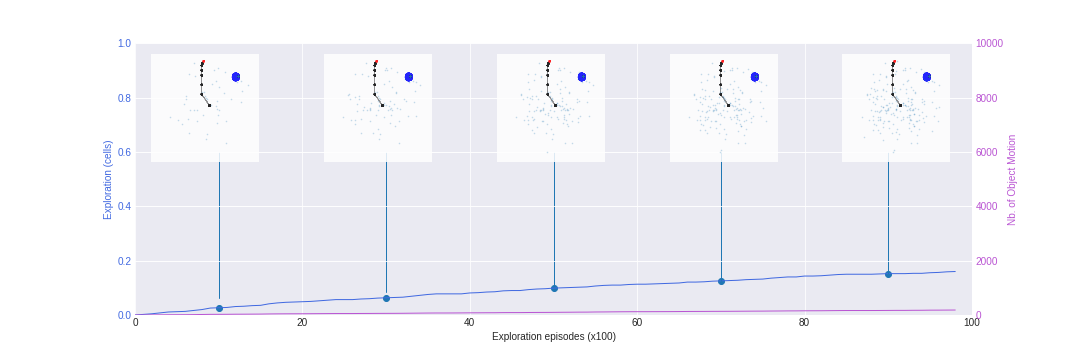}
    \caption{Random Parameterization Exploration}
    \end{subfigure}
    \begin{subfigure}{1.\textwidth}
    \includegraphics[width=\textwidth]{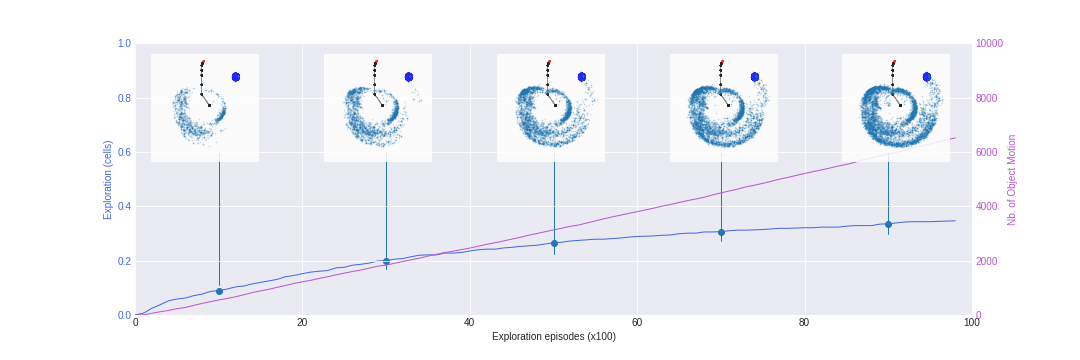}
    \caption{Random Goal Exploration with Engineered Features Representation (RGE-EFR)}
    \end{subfigure}
	\begin{subfigure}{1.\textwidth}
    \includegraphics[width=\textwidth]{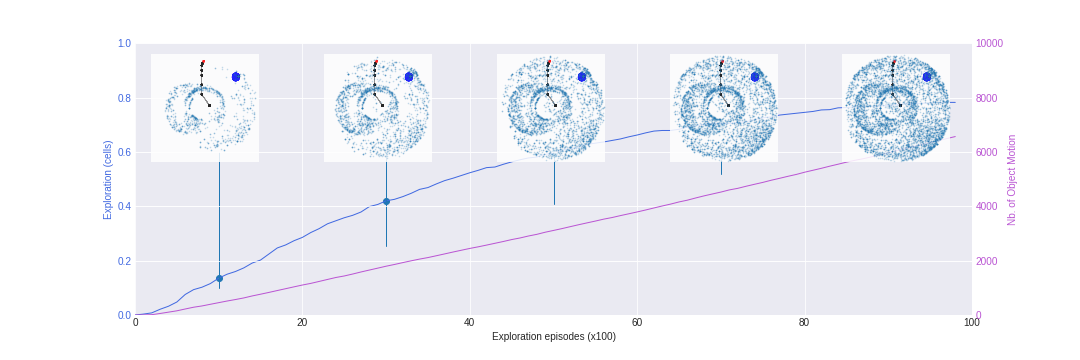}
    \caption{Modular Goal Exploration with Engineered Features Representation (MGE-EFR)}
    \end{subfigure}
    \caption{Examples of achieved observations together with the ratio of covered cells in the \textit{Arm-2-Balls} environment for \textbf{RPE}, \textbf{MGE-EFR} and \textbf{RGE-EFR} exploration algorithms. The number of times the ball was effectively handled is also represented.}
    \label{fig:explo_sprites}
\end{figure}

\begin{figure}
	\centering
    \begin{subfigure}{1.\textwidth}
    \includegraphics[width=\textwidth]{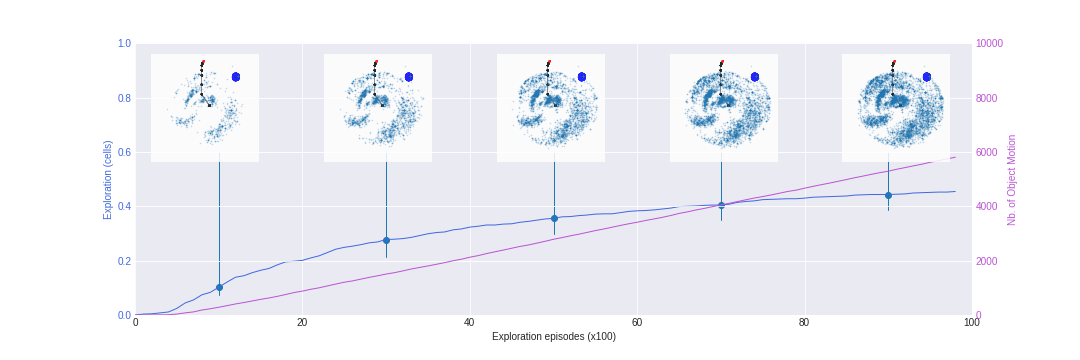}
    \caption{Random Goal Exploration with an entangled representation (VAE) as a goal space (RGE-VAE)}
    \end{subfigure}
	\begin{subfigure}{1.\textwidth}
    \includegraphics[width=\textwidth]{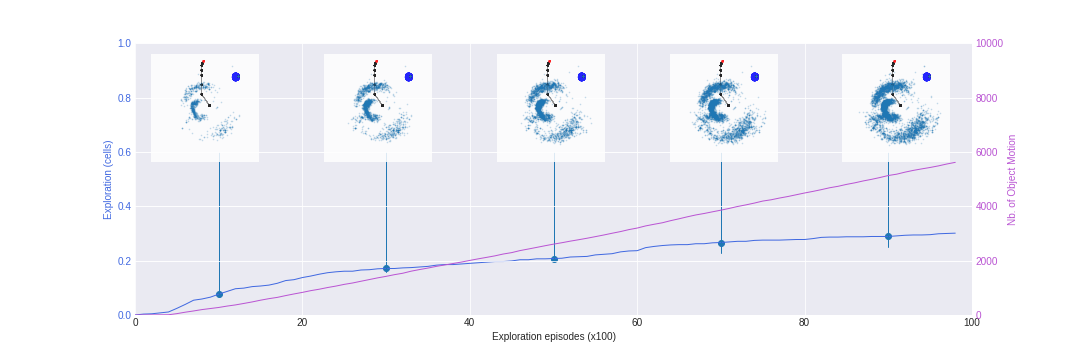}
    \caption{Modular Goal Exploration with an entangled representation (VAE) as a goal space (MGE-VAE)}
    \end{subfigure}
    \begin{subfigure}{1.\textwidth}
    \includegraphics[width=\textwidth]{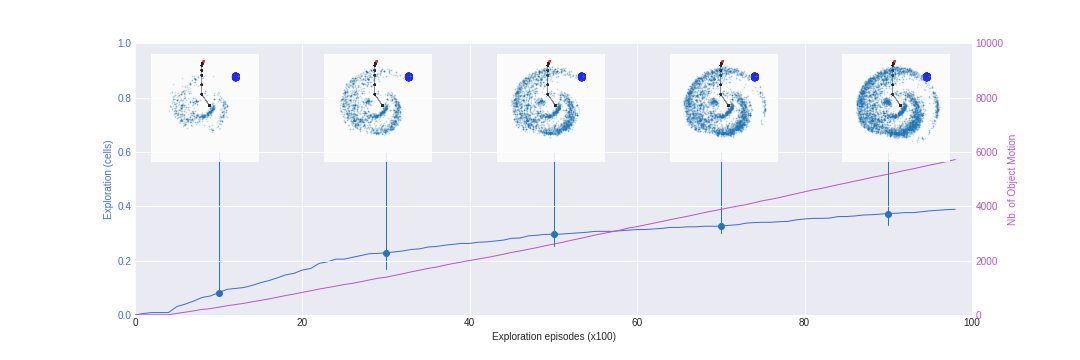}
    \caption{Random Goal Exploration with a disentangled representation ($\beta$VAE) as a goal space (RGE-$\beta$VAE)}
    \end{subfigure}
    \begin{subfigure}{1.\textwidth}
    \includegraphics[width=\textwidth]{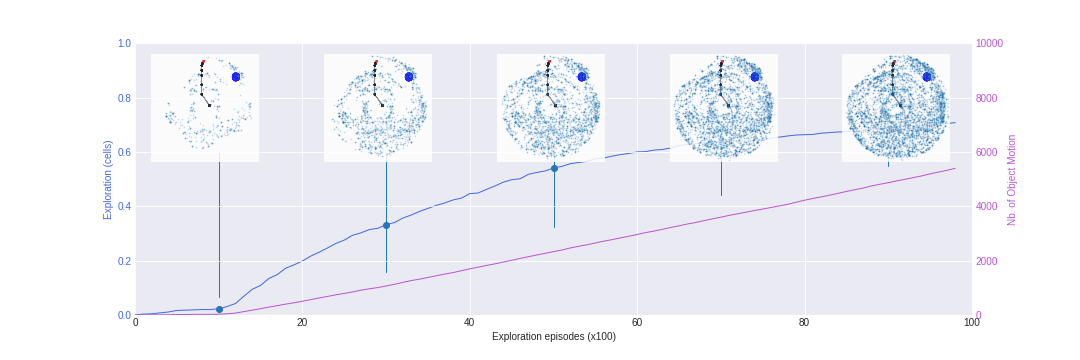}
    \caption{Modular Goal Exploration with a disentangled representation ($\beta$VAE) as a goal space (MGE-$\beta$VAE)}
    \end{subfigure}
    \caption{Examples of achieved observations together with the ratio of covered cells in the \textit{Arm-2-Balls} environment for \textbf{MGE} and \textbf{RGE} exploration algorithms using learned goal spaces (\textbf{VAE} and \textbf{$\beta$VAE}). The number of times the ball was effectively handled is also represented.}
    \label{fig:explo_sprites2}
\end{figure}

Examples of exploration curves obtained with all the exploration algorithms discussed in this paper (\figurename~\ref{fig:explo_sprites} for algorithms with engineered features representation and \figurename~\ref{fig:explo_sprites2} for algorithms with learned goal spaces). It is clear that the random parameterization exploration algorithm fails to produce a wide variety of observations. Although the random goal exploration algorithms perform much better than the random parameterization algorithm, they tend to produce observations that are cluttered in a small region of the space. On the other hand the observations obtained with modular goal exploration algorithms are scattered over all the accessible space, with the exception of the case where the goal space is entangled (VAE).

\end{document}